\documentclass{article}
\usepackage{spconf,amsmath,graphicx}

\usepackage[utf8]{inputenc}
\usepackage[table]{xcolor}
\usepackage{xspace}
\usepackage{tabularx}
\usepackage[justification=centering]{subfig}
\usepackage{url}
\usepackage{balance}
\usepackage{booktabs}

\graphicspath{{figs/}}

\newcommand{\fref}{f.paper}

\def\eg{\emph{e.g.\xspace}}

\newcommand{\diftsc}{DIFT-SC\xspace}
\newcommand{\iftsc}{IFT-SC\xspace}

\newcolumntype{Y}{>{\centering\arraybackslash}X}

\newcommand{\segmented}{SEGMENT3D\xspace}

\newdimen\origiwspc
\newdimen\origiwstr
\origiwspc=\fontdimen2\font
\origiwstr=\fontdimen3\font

\title{SEGMENT3D: A Web-based Application for Collaborative Segmentation of 3D Images Used in the Shoot Apical Meristem}
\name{{\parbox[c]{\textwidth}{\centering Thiago V. Spina$^{1,2,*} \qquad $Johannes~Stegmaier$^{2,3}$  \qquad Alexandre X. Falc{\~{a}}o$^{4}$ \qquad Elliot~Meyerowitz$^{5}$ \qquad Alexandre~Cunha$^{2,*}$ \thanks{We are grateful for funding by the S{\~{a}}o Paulo Research Foundation in projects 2016/11853-2, 2015/09446-7, and 2014/12236-1 (TS, AF), CNPq (AF), the Center for Advanced Methods in Biological Image Analsysis at the Beckman Institute (TS, JS, EM, AC), the German Research Foundation DFG in the project MI1315/4-1 (JS), the Howard Hughes Medical Institute (EM) and the Gordon and Betty Moore Foundation through grant GBMF3406 (EM, AC). Correspondence should be addressed to \texttt{thiago.spina@lnls.br,cunha@caltech.edu}.}}}}

\address{
\small{$^{1}$Brazilian Synchrotron Light Laboratory, Brazilian Center for Research in Energy and Materials, Campinas, SP, Brazil}\\
\small{$^{2}$Center for Advanced Methods in Biological Image Analysis, California Institute of Technology, Pasadena, CA, USA}\\
\small{$^{3}$Institute for Applied Computer Science, Karlsruhe Institute of Technology, Karlsruhe, Germany}\\
\small{$^{4}$Institute of Computing, University of Campinas, Campinas, SP, Brazil}\\ 
\small{$^{5}$Division of Biology and Biological Engineering, California Institute of Technology, Pasadena, CA, USA}}
\begin{document}
%
\maketitle
\begin{abstract}

The quantitative analysis of 3D confocal microscopy images of the \emph{shoot apical meristem} helps understanding the growth process of some plants. Cell segmentation in these images is crucial for computational plant analysis and many automated methods have been proposed. However, variations in signal intensity across the image mitigate the effectiveness of those approaches with no easy way for user correction. We propose a web-based collaborative 3D image segmentation application, {\segmented}, to leverage automatic segmentation results. The image is divided into 3D tiles that can be either segmented interactively from scratch or corrected from a pre-existing segmentation. Individual segmentation results per tile are then automatically merged via consensus analysis and then stitched to complete the segmentation for the entire image stack. \segmented is a comprehensive application that can be applied to other 3D imaging modalities and general objects. It also provides an easy way to create supervised data to advance segmentation using machine learning models.
\end{abstract}

\section{Introduction}
\label{sec:intro}

\begin{figure}[t]
  \begin{center}
  \includegraphics[width=0.47\textwidth]{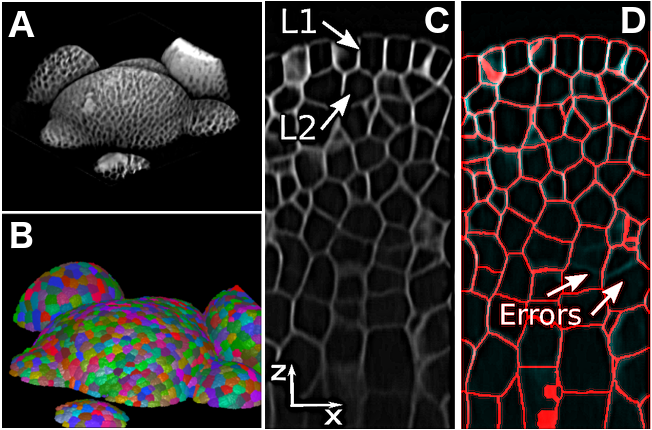}
\caption{A) 3D rendition of a \emph{shoot apical meristem} of \emph{Arabidopsis thaliana} and B) its 3D segmentation showing epithelial cells. C) An $xz$-slice of an enhanced (denoised and sharpened) meristem reveals the weak signal in lower parts. Panel D) shows missing cell boundaries, indicated by arrows, in the automatic 3D segmentation, in red, created by the method in~\cite{Stegmaier:2018:ISBI}.\vspace{-0.2cm}
\noindent\rule{\columnwidth}{0.2mm}
\label{\fref.SAM}
}\vspace{-1.3cm}
\end{center}
\end{figure}

The \emph{shoot apical meristem} is a collection of stem cells of flowering plants responsible for all above ground plant growth
(Figure~\ref{\fref.SAM}). Since most of what humans eat, such as grains and fruits, is derived from the shoot, it is important to investigate the regulation mechanisms that govern meristem growth and consequently the crops that depend on it. Computer simulations done on phantom models with idealized and uniform cell shapes, sizes, and connectivity have recently helped to elucidate cell regulation~\cite{hamant2008}. However, it is desirable to obtain more realistic models derived directly from real data, which require segmentation of thousands of cells in many 3D confocal microscopy images (Figure 1B).

\fontdimen2\font=0.4ex
\fontdimen2\font=\origiwspc

The cells in the meristem are clustered together, forming top layers L1 and L2 (Figure 1C) approximately uniform in cell size and depth. The thickness of the tissue makes it difficult to obtain via confocal microscopy a uniform intensity signal across image layers. This causes the cell membranes at the bottom of the meristem, where the signal is weaker and noisier, to be less evident than those in or near L1 and L2. Many methods have been proposed to segment meristematic cells~\cite{Stegmaier:2018:ISBI,Fernandez10,cunha2010,mkrtchyan2011,federici2012,Reuille15}, but while they work fairly well for L1 and L2, they tend to fail in deeper layers of the meristem (Figure 1D). Deep learning approaches have recently produced superior results in bioimage segmentation, \eg ~\cite{jain2010,raza2017,unet2016}, but those are also incomplete and limited by training data. In the case of the meristem, we lack significant segmented and validated data, specially in deeper layers. We are not aware of a single validated segmentation for a full meristem.

User interaction is thus often required to ensure high quality segmentation results, which are particularly essential when used as training data for deep learning models. However, the development of interactive methods for 3D multi-cell segmentation is challenging. From one aspect, the cognitive overload for a typical user is high when assessing a large number of 3D objects simultaneously. A divide-and-conquer strategy, where small portions of data are annotated one at a time, has shown to reduce such difficulty~\cite{Barreto:2013:COSE,kim2014}. Another difficulty is the interpretation of the 3D image content as projected on a flat screen. It is challenging to communicate visual information and make sound inferences of cell boundaries when only a small portion of the 3D object is displayed. Another challenge is to provide an efficient interactive segmentation algorithm with real-time feedback such that users can quickly assess the results of their editing actions and make sure they are correct. This is crucial to recruit and retain users and guarantee quality. We thus propose a collaborative application, named \segmented, to the 3D image segmentation problem.

\section{Web-based 3D collaborative image segmentation}



We propose \segmented, a collaborative 3D image segmentation application which tackles these difficulties. It is an enabling tool to create good quality segmentation results and generate training data. It is not necessarily a replacement for automatic segmentation algorithms and tools but rather a complement to leverage automatic solutions by providing ways to assess and interactively correct erroneous results. Our application is easily accessible, running on most browsers, aiming to facilitate its dissemination and use by experts and non-experts alike. \segmented borrows ideas from and extends a Collaborative Segmentation application developed earlier to assist segmentation of 2D images of plant cells and other biological images~\cite{Barreto:2013:COSE}. 
\vspace{-0.4cm}




\subsection{Interactive segmentation method}

\segmented provides two main modes of operation: interactive image segmentation and interactive correction of pre-existing results. Either way, the Image Foresting Transform by Seed Competition~\cite{Falcao04a} (\iftsc) is used for delineation. The \iftsc interprets the 3D tile image $I$ as a graph $G=({\cal N},{\cal A}_6)$, in which ${\cal N}$ is the set of voxels of $I$ and ${\cal A}_6$ is an adjacency relation connecting 6-adjacent neighbors. Let $\pi_t=\langle t_1,t_2,\ldots,t \rangle$ represent a simple path in $G$ with terminus at node $t$, such that every $(t_{i},t_{i+1})$ is an arc of ${\cal A}_6$. A function $f(\pi_t)=\max_{i=1,2,\ldots,|\pi_t-1|}\{w(t_{i},t_{i+1})\}$ attributes to every path $\pi_t$ in $G$ a cost given by the maximum weight (gradient) $w(t_{i},t_{i+1})=I(t_{i+1})$ for arc $(t_{i},t_{i+1})\in{\cal A}_6$ along the path, expressing the strength of connectivity that $t$ has to the source of the path. By forcing the paths to be originated from a set ${\cal S}$ of seed nodes, and minimizing an optimum connectivity map $V(t) = \min_{\forall \pi_t \in \Pi_t(G)} \{ f(\pi_t) \}$ for every node $t$ in the graph, the \iftsc partitions $G$ into an \emph{optimum-path forest}, thereby computing the \emph{seeded watershed transform} of the tile. Cells are composed of the optimum-path trees rooted at their seed voxels. The challenge lies in how to properly estimate the seed set, including its optimal location, to render a desired segmentation. Automatic methods rarely produce a faithful seed set.
\vspace{-0.4cm}

\begin{figure}[t]
  \begin{center}
    \includegraphics[width=\columnwidth]{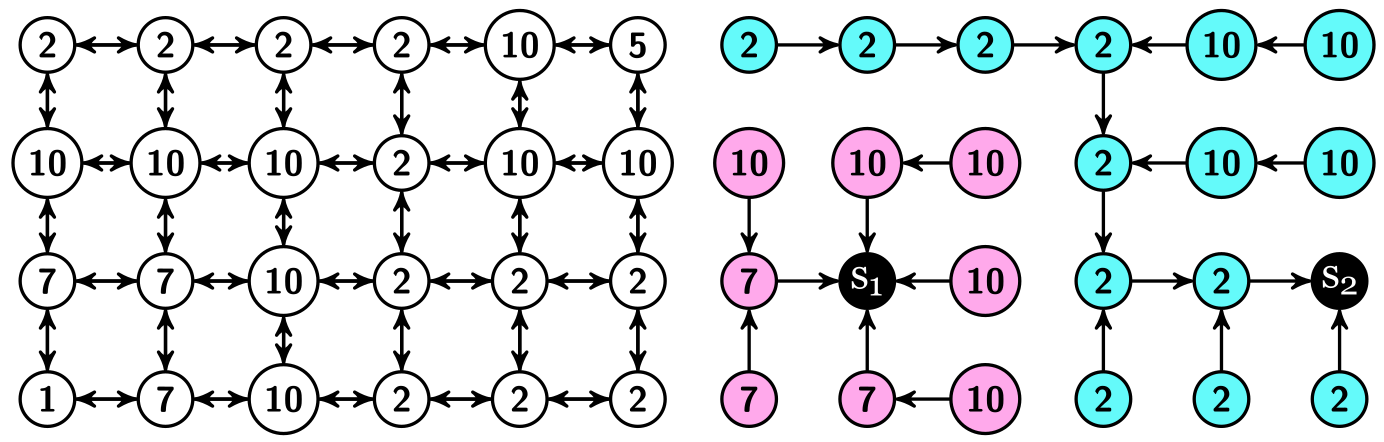}
\caption{\fontdimen2\font=0.4ex 
The figure shows on the left a 4-connectivity image graph with nodes showing the gradient magnitude at the corresponding pixel in the image. In 3D we would have a 6-connectivity graph, ${\cal A}_6$. The gradient ultimately drives the segmentation by seed competition. Two seeds $s_1$ and $s_2$ are added in sequence, shown in black on the right panel. The \iftsc then partitions the graph into an optimum-path forest by forcing the seeds to compete among themselves to conquer the nodes. This partition is the seeded watershed transform~\cite{Falcao04a}.\vspace{-0.2cm}
\noindent\rule{\columnwidth}{0.2mm}
\label{\fref.dift}
}\vspace{-1.3cm}
\end{center}
\end{figure}
\fontdimen2\font=\origiwspc


\subsection{User interaction}

Once a user logs into the system, \segmented searches on the server and feeds an available tile that can be either segmented interactively from scratch or interactively corrected if a pre-segmentation exists. A tile is displayed inside a wireframe box for orientation, and is initially sliced by a clipping plane to show inner portions of the 3D data (Figure~\ref{\fref.interface}). 
The user can then perform segmentation by adding and removing scribbles. The result may be accepted by clicking on {\bf (done)}. If judged too difficult or confusing to segment, the user can opt to {\bf (skip)} and proceed to next tile.
\vspace{-0.4cm}

\begin{figure}[ht]
  \begin{center}
    \includegraphics[width=\columnwidth]{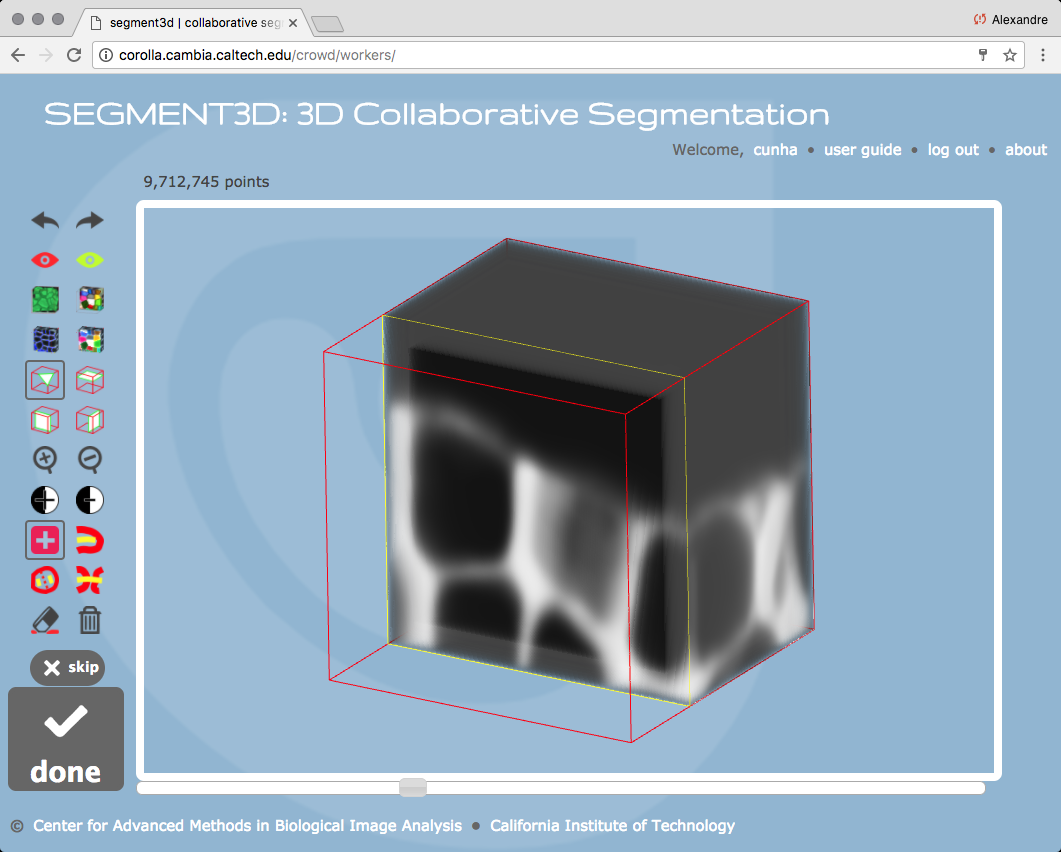}
\caption{\fontdimen2\font=0.4ex
The figure shows our web-based 3D collaborative image
  segmentation application, \segmented, depicting a tile of a 3D image as a self-contained box
  that the user can interactively segment from multiple angles. The tile can be sliced with a clipping plane to view its inner portions at any direction.\vspace{-0.2cm}
  \noindent\rule{\columnwidth}{0.2mm}
\label{\fref.interface}
\fontdimen2\font=\origiwspc
}\vspace{-1.3cm}
\end{center}
\end{figure}

For interactive segmentation, the seeds are simply the set of voxels under the user-drawn scribbles (Figure 4G).
When the user edits scribbles the result is updated locally, in time proportional to the number of voxels in the affected region, using the differential \iftsc~\cite{Falcao04b} (\diftsc). This can be significantly smaller than the total number of voxels in the tile, thus its superior efficiency. We have
simplified user interaction by making each newly added scribble to become the source of a new candidate cell region with a different label. Hence, although all scribbles are displayed with the same color, they each represent a different cell. To correct an existing segmentation, the seed set is
initialized with the geodesic centers of the pre-segmented labels (cell candidates) for \iftsc (Figure 4H).
More sophisticated procedures may also be considered for non-cell objects~\cite{Tavares:2017:ISMM}. Since the user may desire to extend, merge, or split existing cells, the system provides for both interaction modes in total eight operations that can be used when editing scribbles: add, remove, extend, split, merge, undo, redo, and erase all.
\vspace{-0.7cm}

\subsection{Implementation}

We implemented the interface and segmentation methods in JavaScript to run natively on the web browser, aiming to alleviate the server-side programming and slow network traffic. For efficiency, the image is displayed in 3D using the WebGL framework
through the Three.js library~\cite{Threejs:2017}. The backend server code to create tiles, manage user sessions, and store images and segmentation results in a database was written in Python/Django, adapted from~\cite{Barreto:2013:COSE} to 3D.

Ray casting is used  to render the tile volume by treating the image as a 3D
texture, 
inspired by the open source implementation from~\cite{Barbagallo:2017:VOLUMERENDERING} (Figure 4A). 
A challenge we had to consider was how to display many cells at once for segmentation, such that a user can easily decide when cells are properly delineated. We offer 2D and 3D solutions: we can overlay the result directly onto the clipping plane, in a 2D slice fashion (Figure 4B,C), or render the segmentation borders or labels as 3D triangular meshes (Figure~4D,E). In the former, the borders of the segmentation labels and the label image itself are loaded on to WebGL as textures for rendering during ray casting, emulating 2D image view modes.

\begin{figure*}[t]
  \begin{center}
    \includegraphics[width=\textwidth]{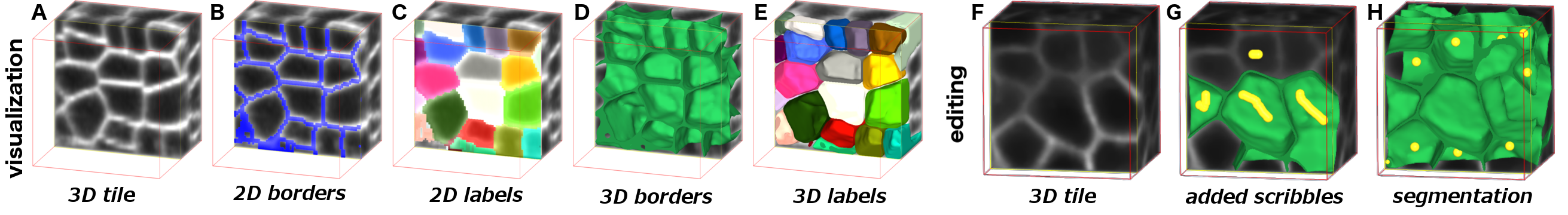}
\caption{\fontdimen2\font=0.4ex 
Users can chose different visualization modes in 2D (panels B,C) and 3D (panels D,E) to evaluate the segmentation after editing. In the interactive segmentation of a 3D tile (F), a user draws scribbles (shown in yellow) over the cells on the clipping plane (G) and the current segmentation is updated and displayed in real time after each added scribble. Each scribble defines a single cell. Some seeds reconstructed at the geodesic centers of the labels for a given pre-segmentation are shown in (H).\vspace{-0.2cm}
\noindent\rule{\textwidth}{0.2mm}
\label{\fref.visu}
\fontdimen2\font=\origiwspc
}\vspace{-1.3cm}
\end{center}
\end{figure*}

Since cells have a well-defined polyhedral shape, the 3D visualization modes facilitate the perception of when errors occur. Due to the confocal acquisition limitations, cell membranes are stronger in the $xy$ focal plane than in $xz$ an $yz$ orthogonal planes. This often causes leaking in the $z$ direction, which are more easily visualized when viewing the 3D meshes generated from the segmentation label or using the 2D borders in the right orientation. We use the Marching Cubes algorithm~\cite{Stemkoski:2013:MARCHINGCUBES} to create cell surface meshes,
whose visual appeal is improved by using a few iterations of Laplacian. The border of the segmentation and of each individual cell are rendered this way (Figure 4D,E).

We keep track of all eight available operations to represent each label by a single scribble. For instance, the user may click on one label and drag the mouse onto another region to perform a label extension. The new scribble is then automatically connected to the scribble representing the first label. Regardless, each operation is always converted to a series of seed additions and removals for the \diftsc. 
\vspace{-0.4cm}


\subsection{Consensus of tile results}

We gather a few results for each tile offered for segmentation. The rationale is that users eventually make mistakes, so by having more than a single segmentation per tile we can potentially automatically eliminate errors by combining many results. We have adopted the STAPLE algorithm~\cite{Warfield:2004:TMI} to create a consensus segmentation for each tile, as it has shown to perform well in practical examples. The method reaches a consensus by using the expectation-maximization algorithm on the borders of the segmentation labels. The \emph{F1-score} of the individual segmentation results are then assessed against the consensus 
to measure the user's accuracy.




\section{Experimental Results}

\newcommand{\tvs}[0]{\texttt{TS}\xspace}
\newcommand{\cunha}[0]{\texttt{AC}\xspace}
\newcommand{\jstegmaier}[0]{\texttt{JS}\xspace}
\newcommand{\falcao}[0]{\texttt{AF}\xspace}
\newcommand{\imagei}[0]{\texttt{Image 1}\xspace}
\newcommand{\imageii}[0]{\texttt{Image 2}\xspace}
\newcommand{\expi}[0]{\texttt{Experiment 1}\xspace}
\newcommand{\expii}[0]{\texttt{Experiment 2}\xspace}
\newcommand{\expiii}[0]{\texttt{Experiment 3}\xspace}

For an initial evaluation of \segmented, four workers (authors TS,JS,AF,AC) used the system and carried out three types of experiments when segmenting a 3D meristem. Our goal was to assess the correctness and usability of the system, in terms of the amount of user effort spent on i) correcting an existing pre-segmentation and ii) interactively segmenting an image from scratch, the two operating modes of \segmented. The first two experiments involved a $128 \times 128 \times 200$ voxel image region from deeper layers of the meristem, where signal is weaker and causes most automated methods to perform poorly. 
It was subdivided into a total of $96$ tiles with size $40\times 40\times40$, with $10\%$ overlap between them and a contextual border of also $10\%$ -- this border facilitates the user's understanding of how the cells are laid out. We segmented the entire region using the method proposed in~\cite{Stegmaier:2018:ISBI} and concluded that only $72$ required interactive correction, after a conservative visual inspection.

\begin{table}[!h]
\caption{User effort and accuracy for all experiments.\label{tab:metrics}\vspace{-0.5cm}}
\begin{center}
{\scriptsize
\begin{tabularx}{\columnwidth}{YYYY}
\bottomrule
\multicolumn{4}{c}{\expi: Pre-segmentation Correction} \\
User & Time (s) & \# interactions & F1-score\\
\midrule
\tvs & $198 \pm 158$ & $1.7 \pm 1.7$ & $0.93\pm0.03$ \\
\cunha & $204 \pm 96$ & $3.3\pm 3.7$ & $0.93\pm0.03$ \\
\jstegmaier & $105\pm 70$ & $1.5\pm 1.5$ & $0.93\pm0.03$ \\
\textbf{Avg}. & $\mathbf{171 \pm 72}$ & $\mathbf{2.2 \pm 2.1}$ & $\mathbf{0.93\pm0.03}$ \\
\toprule
\multicolumn{4}{c}{\expi: Interactive Segmentation}\\
\midrule
\tvs & $185 \pm 53$ & $12.2 \pm 4.0$ & $0.93\pm0.02$\\
\cunha & $335 \pm 125$ & $12.6\pm 4.2$ & $0.93\pm0.02$\\
\jstegmaier & $138\pm 63$ & $10.7\pm 3.8$ & $0.92\pm0.03$\\
Avg. & $219 \pm 68$ & $11.8 \pm 3.4$ & $\mathbf{0.93\pm0.02}$\\

\toprule
\multicolumn{4}{c}{\expii: Pre-segmentation Correction} \\
\midrule
\tvs & $165 \pm 105$ & $2.0 \pm 2.0$ & $0.95 \pm 0.02$\\
\cunha & $203 \pm 111$ & $2.8\pm 2.8$ & $0.95 \pm 0.03$\\
\jstegmaier & $115\pm 101$ & $1.5\pm 1.5$ & $0.95 \pm 0.03$\\
Avg. & $163 \pm 65$ & $2.3 \pm 2.0$ & $0.95 \pm 0.02$\\

\toprule
\multicolumn{4}{c}{\expiii: Pre-segmentation Correction} \\
\midrule
\tvs & $357 \pm 863$ & $4.0 \pm 3.7$ & $0.97 \pm 0.02$\\
\cunha & $319 \pm 133$ & $7.1\pm 4.2$ & $0.97 \pm 0.03$\\
\falcao & $341\pm 409$ & $2.8\pm 6.1$ & $0.95 \pm 0.04$\\
Avg. & $339 \pm 733$ & $4.1 \pm 4.4$ & $0.96 \pm 0.02$\\
\bottomrule
\end{tabularx}
}\vspace{-1.0cm}
\end{center}
\end{table}

In \expi, we compared the two previously stated operating modes i) and ii). We selected $15$ evenly distributed tiles from the $72$ that were considered wrong to be interactively corrected and also interactively segmented from scratch. We measured user effort by evaluating the total amount of the user's time that was required to segment a tile and also the required number of interactions (i.e., operations such as add and delete). Table~\ref{tab:metrics} reveals that interactively segmenting the tiles from scratch took on average $219s$ vs. $171s$ for interactive correction, while the number of interactions was $11.8$ for segmentation from scratch vs. $2.2$ for correction. Hence, interactive correction took $22\%$ less time and $81\%$ less user interaction. However, since each cell required at least one scribble to be segmented, this may indicate that it takes less cognitive effort to segment from scratch than to correct, since the decrease in user interaction was not the same as in the user's time. The extra time was probably making sure scribbles were leading to correct results. Regarding user accuracy, the average F1-score was of $0.93$ for all users, w.r.t. the merger of their results. This value may be explained by the fact that the tiles were selected from deeper layers of the meristem, in which it is harder to obtain a user consensus. Figure~\ref{fig:results} depicts the final result obtained by correcting the pre-segmentation.

In \expii, we asked the users to correct the remaining $57$ tiles. Our goal was 
to verify if the patterns observed in the first experiment were the same. From Table~\ref{tab:metrics}, it took on average $163s$ to make corrections in each tile, consuming about $2.3$ interactions, similarly to before. 
In contrast, user accuracy was higher, since cells from upper layers were present. \vspace{-0.4cm}

\begin{figure}[h]
\begin{center}
\includegraphics[width=\columnwidth]{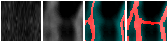}
\caption{\fontdimen2\font=0.4ex
\emph{Left to right}: original tile seen on $xz$ plane, enhanced version for visualization, incorrect pre-segmentation, and correction obtained after the STAPLE consensus for three individual user results.
\label{fig:results}
\vspace{-0.2cm}
\noindent\rule{\columnwidth}{0.2mm}
\fontdimen2\font=\origiwspc
}\vspace{-0.7cm}
\end{center}
\end{figure}

In the final experiment, three workers assessed and corrected $200$ automatically segmented tiles~\cite{Stegmaier:2018:ISBI} with size $60\times60\times60$, from an image region with size $256\times256\times508$. This region was selected to encompass cells ranging from the upper layers L1 and L2 to the bottom of the meristem. The larger size of the tiles doubled the amount of time required for segmentation, on average $339$ seconds, as well as the number of interactions, on average $4.1$, w.r.t. \expii. The accuracy also increased, once again due to the presence of cells from the upper layers with more clearly defined walls. Note that not all users segmented all $200$ tiles, so the results are weighted averages.


\section{Conclusions}

We have presented \segmented, a web-based 3D collaborative image segmentation tool. \segmented 
was designed for cell segmentation, specifically for those of the shoot apical meristem, and can be used both to correct the result of automated methods and to segment an image from scratch interactively. 
Future works involve improving visualization of segmentation, testing the tool with more users, adapting the interface to support mobile device screens, and devising automatic methods for determining the regions of a pre-segmentation label that must be interactively corrected.


\bibliographystyle{IEEEbib}
\bibliography{refs}

\end{document}